\documentclass{article}

\usepackage[preprint]{neurips_2026}
\usepackage[utf8]{inputenc}
\usepackage[T1]{fontenc}
\usepackage{hyperref}
\usepackage{url}
\usepackage{booktabs}
\usepackage{amsfonts}
\usepackage{nicefrac}
\usepackage{microtype}
\usepackage[table]{xcolor}
\usepackage{graphicx}
\usepackage{amsmath, amssymb, bm}
\usepackage{bbm}
\usepackage{multirow}
\usepackage{cleveref}
\usepackage[normalem]{ulem}

\usepackage{xspace}
\newcommand{\ours}{\textsc{CondVLN}\xspace}

\definecolor{best}{RGB}{255,200,200}
\definecolor{second}{RGB}{255,230,170}
\definecolor{third}{RGB}{255,255,190}

\title{\textit{If, Then, Otherwise}: Diagnosing Conditional Branching in Vision-Language Navigation}

\author{
\textbf{Seoyoung Lee}\textsuperscript{1} \quad
\textbf{Neel P. Bhatt}\textsuperscript{1} \quad
\textbf{Pranay Samineni}\textsuperscript{1} \quad
\textbf{Cong Liu}\textsuperscript{2} \\
\textbf{SP Sharan}\textsuperscript{1} \quad
\textbf{Timothy Barclay}\textsuperscript{2} \quad
\textbf{Gregory M. Wagner}\textsuperscript{2} \quad
\textbf{Daniel Milan}\textsuperscript{1} \\
\textbf{Sandeep Chinchali}\textsuperscript{1} \quad
\textbf{Ufuk Topcu}\textsuperscript{1} \quad
\textbf{Atlas Wang}\textsuperscript{1} \\
\normalfont \textsuperscript{1}The University of Texas at Austin, \textsuperscript{2}Collins Aerospace \\
}

\begin{document}

\maketitle

\begin{abstract}
\label{sec:abstract}
Vision-language navigation agents are typically evaluated on their ability to follow route-like instructions toward a fixed goal. Real navigation instructions, however, often depend on the observed state of the environment: \textbf{if} a condition holds, \textbf{then} follow one path, \textbf{otherwise} take another. Such instructions require an agent to evaluate scene evidence, select the correct logical branch, and execute the corresponding navigation behavior. Existing evaluations provide limited control over conditional branch execution, making it difficult to determine whether agents fail because of perception, grounding, navigation, or logical decision-making. We introduce \ours, a scene-graph-grounded benchmark for diagnosing conditional branching in vision-language navigation. \ours programmatically generates instructions whose branch conditions are grounded in verifiable 3D scene-graph predicates, with controlled variation in branch depth, dependency chain length, spatial composition, evidence observability, and instruction horizon. The current instantiation contains over $11{,}500$ generated conditional instructions across AI2-THOR, Matterport3D, Gibson, and ReplicaCAD, and evaluates agents using both standard VLN metrics and branch-specific diagnostics: Branch Selection Accuracy and Conditional Success Rate. Evaluating four state-of-the-art VLN agents, including VLN-Zero, NaVid, NaVILA,
and Open-Nav, shows that conditional branching exposes failures that are not captured by standard success rate or path length alone: agents can navigate plausibly while committing to a branch inconsistent with the observed scene condition. Finally, we present a lightweight neurosymbolic branch-selection model that separates condition grounding from navigation execution, improving performance by 2x. \ours provides a reusable testbed for measuring whether embodied agents can not only follow instructions, but follow the right instruction under the right condition.
\end{abstract}

\section{Introduction}
\label{sec:introduction}

\begin{figure}[t]
    \centering
    \includegraphics[width=\linewidth]{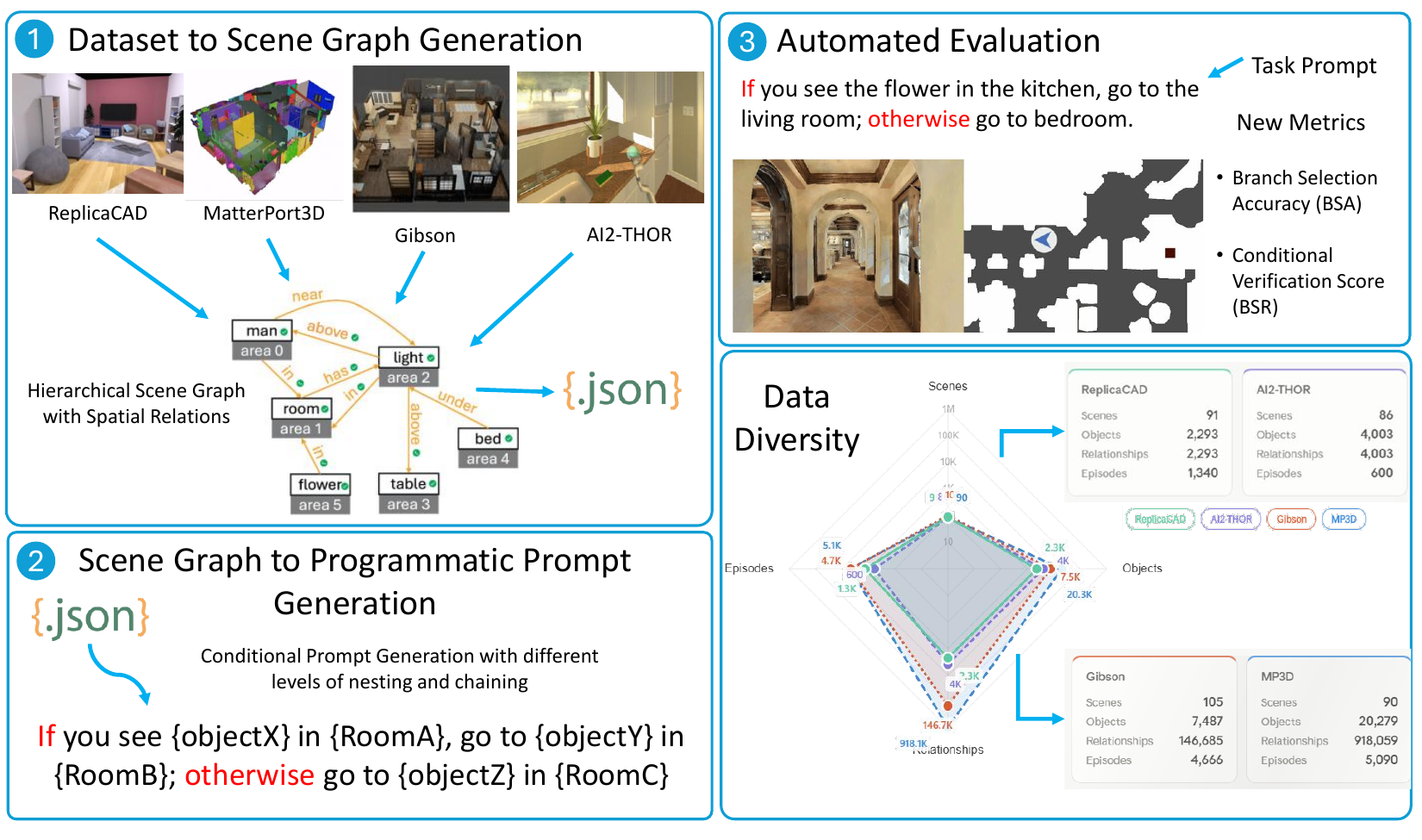}
    \caption{Overview of approach.}
    \label{fig:framework}
\end{figure}

Vision-language navigation (VLN) has emerged as a central testbed for evaluating reasoning capabilities of embodied AI, requiring agents to ground natural-language instructions into sequential decisions in 3D environments. Existing benchmarks such as R2R \citep{r2r}, RxR \citep{rxr}, AI2-THOR \citep{ai2thor}, Gibson \citep{gibson}, ReplicaCAD \citep{habitat}, and Habitat-based extensions \citep{habitat} have significantly advanced multimodal semantic and spatial perception. However, the dominant focus of these benchmarks remains goal-directed navigation, where evaluation focuses on whether agents can reach target locations following instruction trajectories, with limited emphasis on conditional reasoning, branching behavior, and temporal dependency tracking.

In real-world embodied settings, navigation agents must frequently make decisions conditioned on environment state, such as: \emph{If there's a coffee table in the kitchen, head to the stool north of the dishwasher in the kitchen, otherwise make your way to the light below the sink in the bathroom}. These instructions require agents to jointly perform semantic and spatial grounding while also reasoning over explicit conditional structures that govern action selection. Existing datasets neither explicitly control these conditions nor systematically use them for evaluation, limiting controllability over reasoning complexity.

To address this gap, we introduce \ours, a benchmark for conditional reasoning in vision--language navigation that makes conditional reasoning an explicit and controllable evaluation axis. Starting from hierarchical 3D scene graphs that encode visually grounded semantic-spatial relations extracted from embodied environments, we programmatically generate navigation tasks with fine-grained control over logical complexity, including branch depth, dependency chain length, cross-room composition, and instruction horizon. This enables systematic evaluation of how well agents integrate perceptual grounding along with symbolic decision-making. A key property of \ours is that all conditional logic is grounded in explicit scene-graph predicates over object--object and object-room spatial relations, ensuring that branching decisions are semantically verifiable and unambiguous. This design enables precise attribution of failures to perceptual grounding, instruction understanding, conditional reasoning, or execution errors.

\ours is fully compatible with existing VLN ecosystems (e.g., VLN-CE \citep{krantz2020beyond}), enabling evaluation across existing simulators and models without significant infrastructure changes. Beyond standard success metrics, we introduce structured reasoning-specific metrics and evaluation axes for diagnosing conditional navigation failures. By varying instruction length, logical depth, conditional branching structure, and scene diversity, \ours enables fine-grained diagnostic analysis of agent behavior.

Empirical evaluation with state-of-the-art VLN models, including zero-shot instruction-following agents, reveals substantial degradation under conditional branching, highlighting a key limitation of current approaches: the lack of explicit conditional reasoning over structured decision processes.

We further demonstrate the utility of \ours through a conditional branching baseline that explicitly incorporates structured conditional reasoning over scene-graph predicates, consistently improving performance on branching tasks compared to state-of-the-art models. These results suggest that integrating symbolic reasoning over explicit scene structure with neural visual-language grounding is beneficial for robust embodied navigation.

\begin{enumerate}
    \item We introduce conditional reasoning as an explicit evaluation axis for VLN in addition to semantic and spatial reasoning, enabling a holistic and systematic study of neurosymbolic reasoning in embodied navigation.

    \item We propose a hierarchical-scene-graph-based task generation framework that produces semantically verifiable conditional navigation tasks with controllable complexity over branch depth, dependency chains, and spatial composition.
    
    \item We construct a cross-environment VLN benchmark spanning MP3D, Gibson, AI2-THOR, and ReplicaCAD, enabling evaluation across diverse embodied settings without significant infrastructure changes.
    
    \item We introduce controlled reasoning-specific evaluation metrics that disentangle failures due to conditional reasoning through structured variation of task properties.
    
    \item We demonstrate significant performance degradation in state-of-the-art VLN models under conditional branching, establishing \ours as a benchmark for neurosymbolic reasoning in embodied AI.

    \item We provide a neurosymbolic baseline that incorporates structured conditional reasoning, improving performance on branching tasks by 2x.
\end{enumerate}

\section{Related Works}
\label{sec:related_works}

Recent advancements in Large Vision-Language Models (VLMs), such as NaVid \citep{zhang2024navid}, NavGPT-2 \citep{zhou2024navgpt2}, NaVILA \citep{cheng2024navila}, and OmniVLA \citep{hirose2025omnivla}, utilize various methods to achieve better performance on VLN for standard benchmarks such as R2R-CE and RxR-CE \citep{r2r,rxr}. Zero-shot navigation via foundation models has also been explored by methods such as Open-Nav \citep{qiao2025opennav} and CA-Nav (Constraint-Aware Navigator) \citep{chen2024CANav} as an alternative to expert demonstrations, and has achieved good performance in unconventional environments on easily comprehensible instructions as well. However, standard benchmarks in the field either focus primarily on linear instruction following rather than complex conditional logic, or they only use conditional instructions without any organized and structurally verifiable way of assessing them. Thus, the evaluation of these agents is done in a way that yields limited information on performance on long-horizon systematic tasks with logically complex conditional requirements as a central goal of model performance and not just an auxiliary augmentation. 

In particular, other benchmarks like VLNVerse \citep{lin2025vlnversebenchmarkvisionlanguagenavigation}, and IEDL \citep{taioli_r2r_ie} using datasets including AI2-THOR \citep{ai2thor}, Gibson \citep{gibson}, ReplicaCAD \citep{habitat}, R2R \citep{r2r}, and RxR \citep{rxr} primarily focus on whether agents can reach a singular identifiable goal, and so the style of tasks tested by these benchmarks is not extensible to longer-term tasks with logical complexity. Though there has been some progress in testing navigation over long-horizon, multi-part tasks, such as with CoNavBench \citep{wang2026conavbench}, HA-VLN 2.0 \citep{dong2025havlnbenchmarkhumanawarenavigation}, and SeqWalker \citep{han2026seqwalkersequentialhorizonvisionandlanguagenavigation}, there are few benchmarks or frameworks that appropriately address conditional reasoning, branching behavior, or temporal dependency tracking, all tasks characteristic of long-horizon real-world navigation goals and which \ours is designed around. Even recently developed benchmarks such as NavSpace \citep{yang2025navspacenavigationagentsfollow} and LH-VLN \citep{song2024towards}, which test on conditional aspects to account for more stringent VLN requirements, do not use it as a controlled, diagnostic variable, whereas our approach uses a novel scene-graph-verifiable benchmark that allows for detailed configuration of conditional complexity through instruction branch depth and chain length, as well as branch accuracy metrics designed for conditional instructions specifically.

\section{Benchmark Design}
\label{sec:benchmark}

\ours features a programmatic benchmark generation framework for evaluating neurosymbolic reasoning in VLN, where agents must combine semantic-spatial grounding in embodied environments with symbolic reasoning over explicit logical and conditional structures. The framework consists of two parts: dataset generation and evaluation. 

\subsection{Dataset Generation}
\label{sec:dataset_generation}

Dataset generation starts from dataset-specific scene hierarchy files (\Cref{sec:source_datasets,sec:scene_hierarchy}) and converts them into symbolic scene graphs with semantic and spatial relationships (\Cref{sec:spatial-semantic}). These graphs are then used to synthesize conditional navigation instructions with controllable logical complexity (\Cref{sec:conditional_instruction}), which are finally converted into VLN-CE-compatible navigation episodes for evaluation in Habitat-based environments (\Cref{sec:vlnce-realization}).

\subsubsection{Source Datasets and Scene Representations}
\label{sec:source_datasets}
\ours is instantiated across multiple indoor 3D environment sources to evaluate whether conditional navigation generalizes across both synthetic and scan-based scenes. We include synthetic or simulator-native environments, such as AI2-THOR~\citep{ai2thor, procthor, robothor} and ReplicaCAD~\citep{habitat}, as well as scan-based environments, such as MP3D~\citep{Matterport3D} and Gibson~\citep{gibson}. These sources differ in scene scale, annotation format, object taxonomy, and geometry representation, making them useful for testing whether a baseline can generalize across heterogeneous embodied datasets.
 
Datasets built on synthetic data are generated procedurally within a simulator or modeling framework, which can reduce annotation noise and provide complete ground-truth metadata. AI2-THOR is a VLN framework that enables agents to ground natural-language instructions in egocentric visual observations and embodied actions within interactive indoor household scenes. ReplicaCAD is a high-fidelity VLN dataset that provides realistic apartment-scale indoor environments to evaluate language-guided navigation and generalization in physically grounded settings. For evaluation, we generate composed scene GLB files and corresponding navigation mesh files from the scene configurations together with the stage and object GLB assets.

We also utilize another category of datasets generated based on scans of real-world physical scenes invoking techniques such as RGB-D scanning or photogrammetry. These spaces offer a more accurate representation of navigating a real-world indoor space. This includes the MP3D dataset, a database built from scanning 90 properties with a LiDAR camera. Each MP3D scene includes 3D semantic annotations for regions and object instances. The other larger scan-based dataset we included is Gibson's database of 3D Spaces. Gibson is a database of 572 scanned scenes composed of entire buildings. Additionally, Gibson includes scene graphs\citep{armeni20193d} for 105 of these scenes in the format of a four-layered graph.

\subsubsection{Scene Hierarchy Input}
\label{sec:scene_hierarchy}

As each environment source stores semantic and geometric scene information differently, we first convert dataset-specific annotations into a common hierarchy representation. We abstract these heterogeneous sources into a unified scene hierarchy that is represented as a collection of rooms, and each room contains object instances with semantic labels and geometric metadata. At minimum, each object is associated with a 3D center location. When available, the hierarchy may also include object size, radius, floor area, or an explicit 3D axis-aligned bounding box (AABB), which are generated based on the given annotations extracted from the datasets. Formally, for each object $o_i$, we denote its semantic label by $\ell_i$ and its 3D center by $\mathbf{c}_i = (x_i, y_i, z_i)$. The coordinate convention follows the simulator frame, where the $XZ$ plane is treated as the horizontal ground plane and the $Y$ axis is vertical. If an object has a size vector $\mathbf{s}_i = (s_i^x, s_i^y, s_i^z)$, we derive AABB from the object center and extent. If an explicit AABB is provided, it is used directly. Optional radii are used as a fallback geometric approximation when bounding boxes are unavailable.
The hierarchy parser produces a room-level symbolic representation containing object nodes, room containment edges, and available geometric metadata. This representation serves as the foundation for extracting object-object spatial relations and generating scene-graph predicates for conditional instruction synthesis.

\subsubsection{Semantic and Spatial Relationship Generation}
\label{sec:spatial-semantic}

The spatial relationship generator converts each room-level object set into a directed relation graph. Each object is represented as a node, and edges encode both canonical geometric relations and thresholded semantic relations. These relations serve as the symbolic link between spatial-semantic scene structure and the logical conditions used for instruction generation.

\paragraph{Distance metric selection.}
For each ordered object pair $(o_i,o_j)$ in the same room, we compute the relative displacement from source object $o_i$ to target object $o_j$ as $\boldsymbol{\delta}_{i \rightarrow j}=\mathbf{c}_j-\mathbf{c}_i=(\delta_x,\delta_y,\delta_z)$, the center distance $d_{ij}^{\mathrm{center}}=\|\boldsymbol{\delta}_{i\rightarrow j}\|_2$, and the horizontal ground-plane distance $h_{ij}=\sqrt{\delta_x^2+\delta_z^2}$. 

Since source datasets vary in geometric detail, the \ours selects the distance used for semantic thresholding with a prioritized fallback:

$d_{ij}^{\mathrm{used}}$ is selected in the order $d_{ij}^{\mathrm{AABB}} \rightarrow d_{ij}^{\mathrm{sphere}} \rightarrow d_{ij}^{\mathrm{center}}$.
Here, $d_{ij}^{\mathrm{AABB}}=\sqrt{\Delta_x^2+\Delta_y^2+\Delta_z^2}$ is the surface-to-surface AABB distance, where $\Delta_a=\max(a_i^{\min}-a_j^{\max},\,a_j^{\min}-a_i^{\max},\,0)$ for each axis $a\in\{x,y,z\}$, and $d_{ij}^{\mathrm{sphere}}=\max(0,d_{ij}^{\mathrm{center}}-(r_i+r_j))$ is the bounding-sphere surface distance. This fallback allows the same generation framework to operate across datasets with heterogeneous metadata. We prioritize AABBs when available because they capture object extent and face-level geometry with less error.

\paragraph{Canonical directional predicates.}
To generate canonical 3D directional relations, we compute the azimuth and elevation of the displacement vector. Azimuth is discretized into eight equal compass sectors: \{\texttt{east}, \texttt{north-east}, \texttt{north}, \texttt{north-west}, \texttt{west}, \texttt{south-west}, \texttt{south}, and \texttt{south-east}\}. Elevation is discretized into \{\texttt{up}, \texttt{level}, and \texttt{down}\}. 
The final canonical predicate composes the azimuth and elevation bins. Level relations are expressed as horizontal predicates such as \texttt{east of}, while non-level relations include both horizontal and vertical components, such as \texttt{east-and-up of}. These canonical predicates provide a compact symbolic description of relative 3D layout.

\paragraph{Semantic relation predicates.}

In addition to canonical geometric relations, we generate semantic predicates that are closer to real-world natural-language navigation instructions. These predicates are emitted using conservative metric thresholds: \texttt{near} and \texttt{far from} are based on $d_{ij}^{\mathrm{used}}$, \texttt{higher than} and \texttt{lower than} are based on vertical displacement $\delta_y$, and \texttt{above} and \texttt{below} require both sufficient vertical displacement and small horizontal separation.
We use fixed thresholds for proximity and vertical relations to ensure reproducibility across generated datasets.

\paragraph{Scene graph construction.}
For each room, we create a directed graph containing room-object containment edges and object-object relation edges. Each object is assigned a unique identifier and stores its semantic label, center location, and any available geometric metadata. Every ordered object pair includes a canonical directional edge and any semantic edges whose thresholds are satisfied.
Each relation edge stores the subject object, predicate, target object, predicate type, and the metrics used to generate the predicate, including center distance, sphere distance, AABB distance, selected distance type, azimuth bin, and elevation bin. This makes each symbolic relation in the final scene graph traceable to the underlying metric computation.

\subsubsection{Conditional Instruction Synthesis}
\label{sec:conditional_instruction}

    Building on the constructed scene graphs, we synthesize conditional navigation instructions that require agents to reason over object presence and spatial context, and ultimately combine spatial-semantic grounding with symbolic branch execution. Each instruction is generated by grounding the previously generated symbolic predicates from the scene graph into natural-language conditions and navigation actions. They form logical structures such as \textit{conditional branches}, \textit{nested decisions}, and \textit{multi-step chains}.
    As every condition and target is grounded in real object instances and relationships from the scene graph, the generated instructions remain consistent with the underlying 3D environment and the truth value of each condition is known at generation time.

    \paragraph{Object sampling strategy.}
        We construct conditional instructions using a structured object sampling process over the scene graph. Objects are sampled to serve two roles: \textit{valid} references, corresponding to objects that exist in the room, and \textit{invalid} condition references, corresponding to objects that are absent from a given room. Invalid conditions are generated by selecting object categories that exist elsewhere in the scene but not in the reference room, enabling controlled false branches.
        To ensure clarity, objects with identical semantic labels within a room (e.g., multiple ``lamps'' in the room) are disambiguated using spatial qualifiers derived from the scene graph (e.g., ``lamp near the bed''), and objects that cannot be uniquely identified are excluded. This guarantees that all sampled objects are uniquely referable and grounded without ambiguity.

    \paragraph{Instruction templates and complexity.}
        We generate conditional navigation instructions using a structured set of templates that map scene graph predicates into logical forms. At the core, each instruction follows a conditional structure of the form \fbox{\texttt{IF condition THEN action ELSE action}}
        where \texttt{condition} corresponds to an object presence query and \texttt{action} corresponds to navigation targets. This base form can be extended either by nesting conditionals (e.g., \texttt{IF A THEN (IF B THEN x ELSE y) ELSE z}) or by chaining multiple branches (e.g., \texttt{IF A THEN x ELSE IF B THEN y ELSE z}).
        To systematically control reasoning difficulty, we define a set of instruction categories that vary along two axes: \textit{logical depth}, corresponding to the level of nesting, and \textit{chain length}, corresponding to the number of sequential conditional branches. As shown in \Cref{tab:complexity-taxonomy}, these categories allow us to evaluate navigation agents under progressively more complex decision-making scenarios; they range from single-step condition evaluation to multi-step branching and hierarchical reasoning.

        \begin{table}[t]
            \centering
            \caption{Taxonomy of conditional navigation instruction complexity. 
            }
            \begin{tabular}{lccc}
                \toprule
                Category & Depth (nesting) & Chain Length (branches) & Description \\
                \midrule
                Simple & 1 & 1 & Single IF-ELSE \\
                Nested & 2 & 1 & IF within IF \\
                Deep Nested & 3 & 1 & Multi-level nesting \\
                Chain & 1 & 2 & IF / ELSE IF \\
                Long Chain & 1 & 3 & Multi-branch chain \\
                Nested Chain & 2 & 2 & Nested + chained \\
                \bottomrule
            \end{tabular}
            \label{tab:complexity-taxonomy}
        \end{table}

\subsubsection{VLN-CE Dataset Realization}
\label{sec:vlnce-realization}

To make \ours directly usable with existing VLN evaluation frameworks, we convert generated conditional instructions into Habitat-compatible VLN episodes. For each task, we extract start and goal coordinates associated with the selected branch targets from the scene graph and transform coordinates from dataset-native conventions into the Habitat coordinate frame. A Habitat-Sim pathfinder is then used to compute geodesic shortest paths between start and goal states. Episodes with failed pathfinding are filtered out, and the remaining trajectories are stored as reference paths together with geodesic distances, enabling standard VLN metrics.
Each output episode follows the R2R/VLN-CE JSON format, including the scene identifier, start and goal positions, rotation, instruction text, reference path, and geodesic metadata. This step preserves the symbolic conditional structure of each generated instruction while packaging it as a standard navigation episode. As a result, \ours can be evaluated using existing frameworks (e.g., VLN-CE, VLN-Zero, NaVid) by replacing the dataset paths in the corresponding configuration files without significant simulator modifications.
\Cref{fig:framework} summarizes the resulting benchmark coverage.

\subsection{Evaluation Metrics}
\label{sec:evaluation_metrics}

\ours is designed not only as a dataset, but also as a controlled evaluation suite for diagnosing reasoning failures in embodied navigation. Standard VLN metrics, such as success rate (SR), oracle success rate (OSR), distance to goal, and success weighted by path length (SPL), measure whether an agent reaches a target efficiently. However, they do not distinguish failures in visual navigation from failures in conditional reasoning or symbolic branch selection. We therefore introduce reasoning-specific metrics for conditional prompts: branch selection accuracy (BSA) and conditional success rate (CSR). BSA captures branch-consistent progress without requiring final task success, while CSR further requires both complete branch following and standard Habitat success.

\paragraph{Branch selection accuracy.}
Branch Selection Accuracy (BSA) measures how much of the correct conditional branch the agent follows. Let $G_i=(g_{i,1},\ldots,g_{i,m_i})$ denote the ordered subgoals associated with that branch for episode $i$, and let $k_i$ be the length of the longest prefix of $G_i$ reached in order within Habitat's success radius $\tau$. For episodes with $m_i>0$, we define $\mathrm{BSA}_i=k_i/m_i$, allowing fractional credit for partial branch progress. For episodes with no annotated subgoals, branch progress is not separately observable from final-goal success, so successful no-subgoal runs are assigned $\mathrm{BSA}_i=1$, while unsuccessful no-subgoal runs are excluded. Let $\mathcal{I}$ denote the resulting set of included episodes. We report $\mathrm{BSA}=|\mathcal{I}|^{-1}\sum_{i\in\mathcal{I}}\mathrm{BSA}_i$.

\paragraph{Conditional success rate.}
Conditional Success Rate (CSR) measures strict conditional task completion by requiring both complete branch following and final navigation success. Using the same amended inclusion set $\mathcal{I}$ as BSA, we define $\mathrm{CSR}_i=\mathbbm{1}[\mathrm{BSA}_i=1 \land \mathrm{Success}_i=1]$, where $\mathrm{BSA}_i$ is the reported branch score after the no-subgoal convention and $\mathrm{Success}_i$ is the standard VLN-CE/Habitat success metric. Thus, nonempty-subgoal episodes require all ordered branch subgoals to be completed and the agent to succeed at the final goal, while successful no-subgoal episodes are assigned $\mathrm{CSR}_i=1$ and unsuccessful no-subgoal episodes are excluded. We report $\mathrm{CSR}=|\mathcal{I}|^{-1}\sum_{i\in\mathcal{I}}\mathrm{CSR}_i$.

\section{Neurosymbolic Branch-Selection Oracle Model}
\label{sec:neurosymbolic_baseline}

\ours is designed to distinguish failures in conditional reasoning from failures in downstream navigation. To support this diagnostic goal, we introduce a lightweight \textit{oracle} baseline that resolves the correct conditional branch using symbolic information recorded during benchmark construction, and then relays a linearized navigation instruction to an unchanged VLN agent. The baseline serves as a controlled diagnostic probe rather than a new navigation architecture: \textit{what happens if the logical branch of a conditional instruction is selected correctly before the instruction is passed to a standard VLN agent}? 
It estimates the effect of resolving symbolic branch selection while keeping downstream perception, goal grounding, and navigation execution unchanged.
We instantiate this oracle on top of VLN-Zero because its scene-graph-based reasoning is closely aligned with the symbolic scene structure used by \ours. This makes it a natural testbed for isolating the effect of symbolic branch resolution. In contrast, methods such as Open-Nav perform spatio-temporal reasoning primarily in image space, making the symbolic oracle less directly aligned with their internal reasoning representation.

\paragraph{Oracle branch selector.}
Given a generated conditional instruction, the oracle branch selector uses symbolic metadata recorded at dataset construction time to determine the correct branch. Since each instruction is programmatically generated based on a scene-graph, the truth value of each condition and the corresponding branch goal are known by construction. Rather than re-evaluating the scene graph during simulator evaluation, we use the recorded structured form of the instruction, together with the selected goal and any ordered subgoal annotations, to rewrite the agent-facing prompt into a linear navigation instruction.

Concretely, an instruction of the form
``if condition $c$ holds, go to $g_{\mathrm{true}}$; otherwise, go to $g_{\mathrm{false}}$''
is rewritten as ``go to $g_b$,'' where $b\in\{\mathrm{true},\mathrm{false}\}$ is the ground-truth branch determined by the scene graph and $g_b$ is the corresponding branch goal.
For nested and chained conditionals, this procedure is applied recursively until a leaf goal is reached.
When ordered subgoals $g_1,\ldots,g_m$ are available, the oracle instruction instead linearizes them as an ordered waypoint chain before the terminal goal, e.g., ``first go to $g_1$, then go to $g_2$, and finally go to $g_b$.''
Thus, the resulting ``oracle-linear'' version of each episode instruction removes explicit if-then-else structure from the agent-facing instruction while preserving the selected goal and any subgoal structure already encoded for that episode.

\paragraph{Evaluation role and implementation.}
As a diagnostic probe, the oracle baseline estimates how well a fixed VLN agent performs when the conditional branch has been resolved by construction. We implement it as a deterministic dataset transformation: only the agent-facing instruction is rewritten into its oracle-linear form, while the underlying navigation episode, including goals, subgoals, reference paths, and geodesic metadata, is kept unchanged. Therefore, differences between the original conditional setting and the oracle setting primarily reflect the removal of explicit branch-selection difficulty rather than changes to the simulator, navigation targets, or downstream model.

\section{Experimental Results}
\label{sec:experiments}

\paragraph{Experiment overview.}
Our experiments test whether current VLN agents can handle the logical conditional structure introduced by \ours. In particular, we ask three questions: (i) which agents perform best across environments, (ii) how performance changes as conditional instruction structure becomes more complex, and (iii) whether failures are driven primarily by nested logical depth or by sequential chain length. The benchmark coverage and dataset statistics are summarized in \Cref{fig:framework}.
Here, we focus on model evaluation using standard VLN metrics (SR and SPL) together with reasoning-specific metrics introduced in \Cref{sec:evaluation_metrics} (BSA and CSR).

\subsection{Experimental Setup}
\label{sec:exp_setup}

\paragraph{Datasets and Environments.}
We evaluate \ours across a diverse set of embodied navigation environments spanning both continuous and discretized settings. Specifically, we instantiate our benchmark on AI2-THOR, Gibson, ReplicaCAD, and Matterport3D (MP3D) as shown in \Cref{sec:source_datasets} and \Cref{fig:framework}.

\paragraph{Evaluated Agents.}
We evaluate four state-of-the-art VLN agents representing diverse modeling frameworks: VLN-Zero, NaVid, NaVILA, and Open-Nav. These agents differ in their use of pretrained vision-language models, memory mechanisms, and planning strategies, providing a broad view of how existing VLN approaches handle conditional reasoning. Unless explicitly required by the original model setup, all agents are evaluated without retraining to preserve comparability with reported baselines.

\paragraph{Implementation Details.}
All experiments are conducted using a unified evaluation interface built on VLN-CE-compatible infrastructure to ensure consistency across environments. Episodes are executed with a fixed action budget proportional to instruction length. The success-distance threshold is set to 0.5 m for single-room scenes (i.e., ReplicaCAD) and 1.0 m for other datasets with multi-room scenes.
For conditional tasks, agents are not given explicit branch annotations beyond the natural-language instruction. They must implicitly infer the branch condition, evaluate the relevant scene evidence, and execute the corresponding navigation behavior.

\subsection{Main Results}
\label{sec:main_results}

We present the main benchmark results across all four datasets and instruction-complexity categories in \Cref{tab:main_results}. Overall, conditional navigation designed with explicit reasoning or structured representations remains challenging for existing VLN agents. General-purpose, non-custom VLN agents such as NaVid and NaVILA struggle across all datasets, exhibiting low and unstable CSR. This indicates that current training paradigms do not generalize well to unseen environments or to structured conditional decision-making in 3D scenes.

In contrast, Open-Nav and VLN-Zero demonstrate substantially stronger performance, likely due to more explicit reasoning structures. Open-Nav benefits from zero-shot LLM-based reasoning combined with spatio-temporal chain-of-thought planning, while VLN-Zero leverages explicit scene-graph structure to better capture environmental relationships. These mechanisms improve robustness under conditional instructions, particularly in settings requiring compositional reasoning.

The Oracle model shows a different behavior pattern: while it does not consistently dominate in the aggregated results, its advantages emerge more clearly under increasing complexity (as further shown in \Cref{tab:depth_chain_scaling}). In simpler settings, higher success rates across models lead to relatively inflated BSA and CSR, which partially obscures the benefits of explicit symbolic reasoning in the overall mean. However, Oracle-style decomposition becomes increasingly effective as instruction complexity grows, suggesting that its strengths are most pronounced in non-trivial conditional scenarios.

Overall, these results show that standard VLN metrics alone are insufficient for evaluating conditional navigation. Agents may achieve nontrivial SR while still obtaining much lower conditional metrics, indicating that reaching a plausible goal does not necessarily imply correct branch execution.

\subsection{Depth and Chain Scaling}
\label{sec:depth_chain_scaling}

\Cref{tab:depth_chain_scaling} further decomposes instruction complexity along two controlled axes: logical depth and chain length. This analysis isolates whether performance degradation is driven primarily by hierarchical conditional nesting or by long-horizon sequential dependencies. Overall, increasing either dimension leads to consistent degradation in BSA and especially CSR, confirming that conditional VLN becomes progressively harder as reasoning complexity increases.

NaVid and NaVILA remain weak across all configurations, and their performance degrades further as depth and chain length increase, particularly when moving from $d{=}2$ to $d{=}3$, from intermediate to more complex settings. This highlights their limited ability to handle compositional or multi-step conditional reasoning, and confirms that the benchmark effectively stresses instruction complexity.
In contrast, Open-Nav maintains relatively stable performance across both sweeps due to its LLM-based reasoning and spatial grounding, though it still exhibits noticeable degradation under higher complexity. VLN-Zero also remains competitive, but shows clearer sensitivity to increasing depth and chain length, reflecting the limitations of scene-graph-based reasoning under extended conditional dependencies.

Most importantly, Oracle model consistently outperforms VLN-Zero across all controlled settings, particularly under deeper and longer conditional structures. It also surpasses Open-Nav by twofold in challenging regimes such as $d{=}3,\ell{=}1$ and $d{=}1,\ell{=}2$, demonstrating the effectiveness of explicit symbolic decomposition when reasoning complexity increases. Overall, these results confirm that while learned reasoning improves robustness, explicit symbolic structure remains critical for reliable conditional decision-making in complex VLN settings.

\begin{table}
  \caption{ 
  Main results on \ours averaged across all instruction categories. We report branch selection accuracy (BSA) and conditional success rate (CSR).}
  \label{tab:main_results}
  \centering
  \resizebox{\textwidth}{!}{
  \begin{tabular}{lcccccccccccccccc}
    \toprule
    \multirow{2}{*}{Model}
    & \multicolumn{4}{c}{ReplicaCAD}
    & \multicolumn{4}{c}{AI2-THOR}
    & \multicolumn{4}{c}{Gibson}
    & \multicolumn{4}{c}{Matterport3D} \\
    \cmidrule(lr){2-5}
    \cmidrule(lr){6-9}
    \cmidrule(lr){10-13}
    \cmidrule(lr){14-17}
    & SR & SPL & BSA & CSR
    & SR & SPL & BSA & CSR
    & SR & SPL & BSA & CSR
    & SR & SPL & BSA & CSR \\
    \midrule

    NaVid    
    & 16.3 & 16.0 & 15.6 & 9.20
    & 18.4 & 16.3 & 12.8 & 11.2
    & 1.0 & 0.9 & 1.5 & 0.0
    & 5.1 & 2.58 & 2.4 & 0 \\

    NaVILA  
    & 0.0 & 0.0 & \cellcolor{third}17.3 & 0.0
    & 0.0 & 0.0 & 16.5 & 0.0
    & 0.0 & 0.0 & 3.0 & 0.0
    & 2.0 & 1.88 & 7.0 & 0.0 \\

    Open-Nav 
    & \cellcolor{best}29.0 & \cellcolor{second}11.5 & \cellcolor{best}39.2 & \cellcolor{best}33.3
    & \cellcolor{best}54.0 & \cellcolor{second}22.5 & \cellcolor{second}31.8 & \cellcolor{best}29.9
    & \cellcolor{best}12.0 & \cellcolor{best}7.0 & \cellcolor{second}1.6 & \cellcolor{third}1.1
    & \cellcolor{second}18.0 & \cellcolor{second}8.0 & \cellcolor{second}11.4 & \cellcolor{second}7.1 \\

    VLN-Zero 
    & \cellcolor{second}28.0 & \cellcolor{best}26.6 & \cellcolor{second}20.3 & \cellcolor{second}14.0
    & \cellcolor{second}47.0 & \cellcolor{third}37.6 & \cellcolor{third}31.5 & \cellcolor{third}21.0
    & \cellcolor{second}8.0 & \cellcolor{second}6.1 & \cellcolor{best}6.8 & \cellcolor{best}2.1
    & \cellcolor{best}19.0 & \cellcolor{best}19.0 & \cellcolor{best}14.1 & \cellcolor{best}9.9 \\

    Oracle 
    & \cellcolor{third}22.0 & \cellcolor{third}22.0 & 15.3 & \cellcolor{third}12.0
    & \cellcolor{third}46.0 & \cellcolor{best}37.7 & \cellcolor{best}34.5 & \cellcolor{best}29.0
    & \cellcolor{third}7.0 & \cellcolor{third}4.9 & \cellcolor{third}6.3 & 2.1
    & \cellcolor{third}16.0 & \cellcolor{third}14.7 & \cellcolor{third}10.1 & \cellcolor{third}5.8 \\

    \bottomrule
  \end{tabular}
  }
\end{table}

\begin{table}[t]
  \caption{Scaling analysis on \ours. The base case with depth $d{=}1$ and chain length $\ell{=}1$ is shared by both the logical-depth and chain-length sweeps and is therefore reported only once.}
  \label{tab:depth_chain_scaling}
  \centering
  \resizebox{\textwidth}{!}{
  \begin{tabular}{lcc cc cc cc cc}
    \toprule
    \multirow{3}{*}{Model}
    & \multicolumn{2}{c}{Base}
    & \multicolumn{4}{c}{Logical Depth Sweep}
    & \multicolumn{4}{c}{Chain Length Sweep} \\
    \cmidrule(lr){2-3}
    \cmidrule(lr){4-7}
    \cmidrule(lr){8-11}
    & \multicolumn{2}{c}{$d{=}1,\ell{=}1$}
    & \multicolumn{2}{c}{$d{=}2,\ell{=}1$}
    & \multicolumn{2}{c}{$d{=}3,\ell{=}1$}
    & \multicolumn{2}{c}{$d{=}1,\ell{=}2$}
    & \multicolumn{2}{c}{$d{=}1,\ell{=}3$} \\
    \cmidrule(lr){2-3}
    \cmidrule(lr){4-5}
    \cmidrule(lr){6-7}
    \cmidrule(lr){8-9}
    \cmidrule(lr){10-11}
    & BSA & CSR
    & BSA & CSR
    & BSA & CSR
    & BSA & CSR
    & BSA & CSR \\
    \midrule
    NaVid & 5.67 & 1.25 & 10.88 & 5.88 & 10.29 & 7.50 & 16.68 & 10.14 & 7.48 & 2.27 \\
    NaVILA & 5.35 & 0.00 & 7.22 & 0.00 & 4.28 & 0.00 & 12.06 & 0.00 & 10.42 & 0.00 \\
    Open-Nav & \cellcolor{best}22.94 & \cellcolor{best}17.21 & \cellcolor{third}17.70 & \cellcolor{best}14.06 & \cellcolor{second}14.59 & \cellcolor{second}12.27 & \cellcolor{third}21.88 & \cellcolor{third}11.51 & \cellcolor{best}13.37 & \cellcolor{best}11.29 \\
    VLN-Zero & \cellcolor{third}17.20 & \cellcolor{third}8.97 & \cellcolor{second}23.30 & \cellcolor{third}11.78 & \cellcolor{third}13.87 & \cellcolor{third}10.08 & \cellcolor{second}25.76 & \cellcolor{second}18.49 & \cellcolor{second}13.36 & \cellcolor{second}9.91 \\
    Oracle & \cellcolor{second}18.60 & \cellcolor{second}13.33 & \cellcolor{best}23.82 & \cellcolor{second}13.99 & \cellcolor{best}22.66 & \cellcolor{best}17.63 & \cellcolor{best}28.23 & \cellcolor{best}24.63 & \cellcolor{third}8.33 & \cellcolor{third}6.47 \\
    \bottomrule
  \end{tabular}
  }
\end{table}

\section{Conclusion and Limitations}
\label{sec:conclusion_limitations}

We introduced \ours, a benchmark for evaluating conditional branching in vision-language navigation. By programmatically generating conditional navigation instructions from grounded scene graphs, \ours enables controlled evaluation of branch reasoning, condition grounding, and navigation execution. The benchmark spans four embodied navigation environments including AI2-THOR, MP3D, Gibson, and ReplicaCAD, providing diverse navigation settings and reasoning complexity. Our experiments show that current VLN agents struggle significantly under conditional branching, revealing failure modes not captured by standard navigation metrics alone. We further demonstrated that incorporating explicit symbolic branch reasoning through a lightweight oracle neurosymbolic baseline improves branch selection and conditional execution performance. Overall, \ours provides a reusable testbed for studying neurosymbolic reasoning and conditional decision-making in embodied navigation.

\ours currently focuses on VLN-CE-compatible indoor navigation datasets, limiting evaluation to existing simulator ecosystems. The benchmark also inherits imperfections from the underlying scene scans and semantic annotations across the four environments, which can affect perceptual grounding and condition verification. In addition, our spatial relation generation relies on fixed geometric heuristics that may not generalize across all scene distributions or environment types. Finally, our neurosymbolic baseline uses oracle scene-graph information for branch reasoning, avoiding the uncertainty faced by real embodied agents. Future work will extend \ours to outdoor and open-world settings, develop more adaptive relation-generation methods, and investigate stronger perception-driven and LLM+symbolic reasoning baselines.

\bibliographystyle{plainnat}
\bibliography{bib/references}

\end{document}